\pdfoutput=1

\documentclass[11pt]{article}

\usepackage{acl}

\usepackage{times}
\usepackage{latexsym}

\usepackage[utf8]{inputenc} 
\usepackage[T1]{fontenc}    
\usepackage{hyperref}       
\usepackage{url}            
\usepackage{booktabs}       
\usepackage{amsfonts}       
\usepackage{nicefrac}       
\usepackage{microtype}      
\usepackage{xcolor}         
\usepackage{graphicx}
\usepackage{multirow}
\usepackage{multicol}
\usepackage[colorinlistoftodos]{todonotes}
\usepackage[nameinlink]{cleveref}

\newcommand{\algfont}[1]{\texttt{#1}}

\usepackage[T1]{fontenc}

\usepackage[utf8]{inputenc}

\makeatletter
\def\blfootnote{\xdef\@thefnmark{}\@footnotetext}
\makeatother

\usepackage{microtype}

\title{Pretrained Models for Multilingual Federated Learning}

\author{Orion Weller*, Marc Marone*, \\
\textbf{Vladimir Braverman, Dawn Lawrie, Benjamin Van Durme} \\ 
Johns Hopkins University \\
\normalsize{\texttt{oweller@cs.jhu.edu,mmarone1@jhu.edu}}
}

\begin{document}
\maketitle
\begin{abstract}
    Since the advent of Federated Learning (FL), research has applied these methods to natural language processing (NLP) tasks.
    Despite a plethora of papers in FL for NLP, no previous works have studied how multilingual text impacts FL algorithms. 
    Furthermore, multilingual text provides an interesting avenue to examine the impact of non-IID text (e.g. different languages) on FL in naturally occurring data.
    We explore three multilingual language tasks, language modeling, machine translation, and text classification using differing federated and non-federated learning algorithms. 
    Our results show that using pretrained models reduces the negative effects of FL, helping them to perform near or better than centralized (no privacy) learning, even when using non-IID partitioning.\footnote{Our code and data are made publicly available at \url{https://github.com/orionw/Multilingual-Federated-Learning}}
\end{abstract}

\section{Introduction}
\blfootnote{* Authors contributed equally}

Federated learning (FL) is a machine learning technique that trains a model across multiple distributed clients holding local data samples, without ever storing client data in a central location \citep{konevcny2016federated,mcmahan2017communication}.
These techniques are appealing for those who wish to learn from data in a privacy-preserving way, without ever transmitting the data off of a client device.
FL becomes essential when data is especially sensitive, as is the case at hospitals, legal firms, financial institutions, or in countries that enact legislation concerning data privacy (such as the EU's GDPR or the US's HIPAA). 

\begin{figure}[t!]
    \centering
    \includegraphics[trim=10 0 0 0,clip,width=0.5\textwidth]{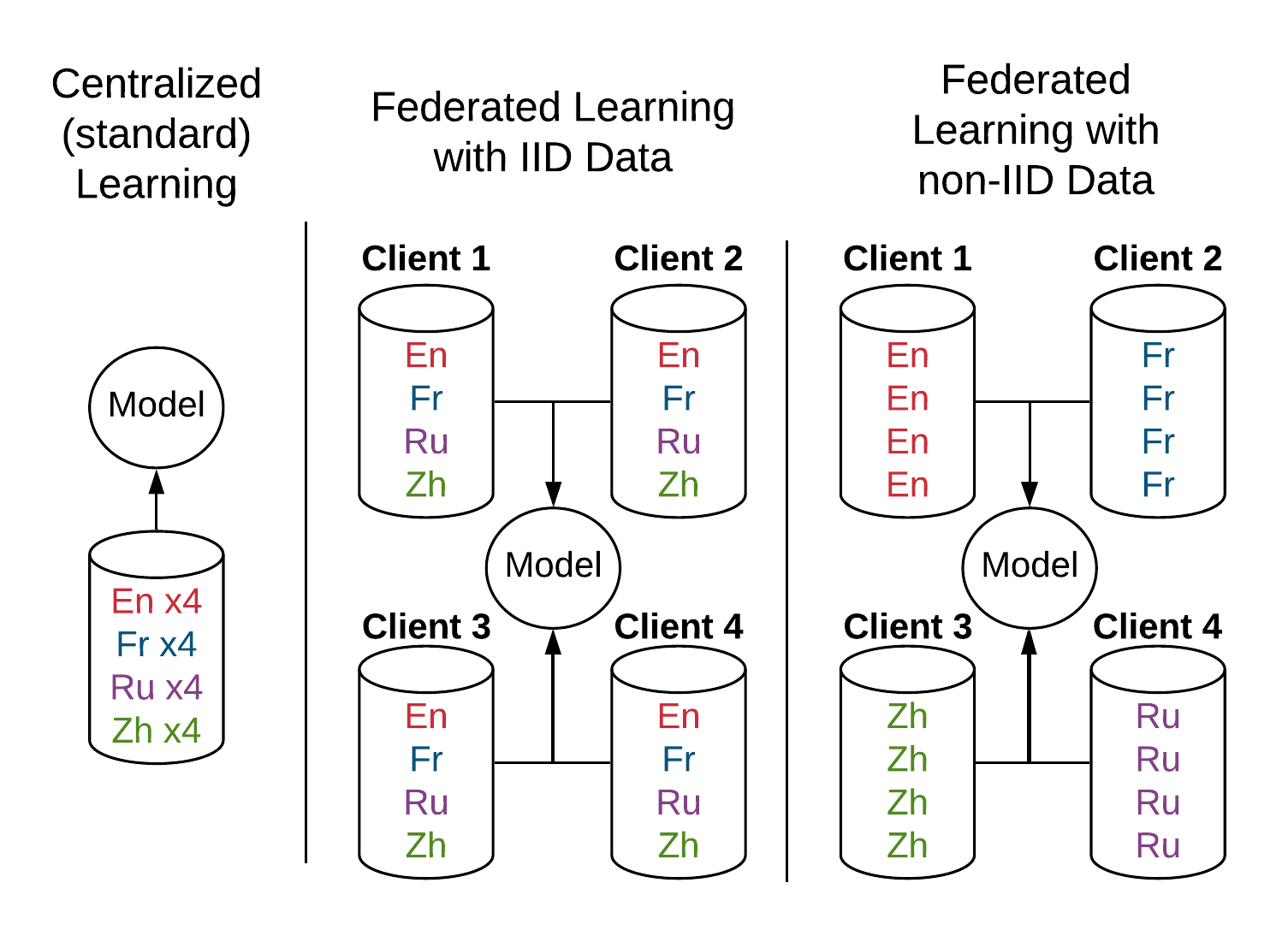}
    \caption{A depiction of different learning strategies with Federated Learning (FL) and multilingual data, with 4 clients and 16 instances from En, Fr, Ru, and Zh in this toy example. Black lines indicate gradient flow. Centralized learning is the standard training method (no privacy), FL with IID data partitions the data into IID data subsets for each client, while FL with non-IID data has the languages separated across clients.}
    \label{fig:task}
\end{figure}

FL has been applied to problems in natural language processing (NLP) since its inception, particularly in use of the language modeling task \citep{yang2018applied,hard2018federated,ramaswamy2019federated,chen2019federated,ji2019learning,stremmel2020pretraining}. Another large area of FL research is focused on analyzing performance when the data is non identically independently distributed (non-IID). In these cases, many works have shown that FL performance is sub-par with respect to centralized learning methods \cite{konevcny2016federated,hard2018federated,fednlp2021}.

Despite the large amount of research in FL for NLP, how different languages impact the FL training process has yet to be explored \citep{liu2021federated}. Furthermore, multilingual FL provides an interesting and natural setting to explore non-IID data, of which different languages are an obvious example.

In this work, we explore multilingual federated learning across three multilingual language tasks and different stages of model pretraining. Our results show that fine-tuning pretrained models with FL can perform similarly to pretrained models fine-tuned with the standard centralized method (the no privacy setting), despite having completely non-IID language partitioned data. This finding shows that pretrained models provide an effective way for practitioners (and consumers) of multilingual data to gain the privacy benefits of FL at little or no cost to the final task performance.

\section{Background and Related Work}
The term \textit{Federated Learning} was first proposed in \citet{mcmahan2017communication}, who applied the \algfont{FederatedAveraging} algorithm to the tasks of language modeling and image classification. 
Since then, much of the theoretical and applied work in FL (e.g. \citet{chen2019federated_b,wu2020fedmed} among many others) has considered language modeling as a key task or benchmark. 

Concurrent with the growing interest in Federated Learning, NLP has rapidly shifted toward the use of pretrained language models (PLMs) (e.g., BERT \citealt{devlin2019bert}; T5 \citealt{raffel2019exploring}; GPT-3 \citealt{brown2020language}).
These PLMs are used for both the core task of next word prediction and as a starting point for learning other downstream NLP tasks. This \textit{pretrain-and-fine-tune} paradigm has since become ubiquitous in modern NLP and has inspired a large and active area of research in model pretraining.
Multilingual versions of these pretrained models have since been developed and are often used with transfer learning techniques to increase performance for tasks where data is limited (e.g. mBERT from \citealt{devlin2019bert}).

The intersection of distributed learning from private data partitions
and PLMs is still a nascent area.
Several works have explored more efficient methods of federated communication with the purpose of enabling these larger NLP models for production situations \cite{Sui2020FedEDFL,Wu2021FedKDCE}.
Our work is orthogonal to these (and could be combined in future work), as we explore the effects of multilingual data on PLM FL, rather than creating methods to enable their use.
Other papers focus on the gap between federated learning performance and centralized performance, evaluating on a wide variety of English NLP tasks \cite{Liu2020FederatedPA,fednlp2021,Chen2021FedMatchFL}.
Although they focus on differential privacy (DP) rather than FL, \citet{li2021large} find that direct PLM training is difficult with standard DP methods, but that fine-tuning PLMs on English data is possible with private learning techniques.
We differ from all these works by studying private learning, specifically FL, for PLMs in the novel multilingual setting. 

\section{Experimental Design}

\begin{figure*}[t]
    \centering
    \includegraphics[trim=10 10 0 0,clip,width=0.48\textwidth]{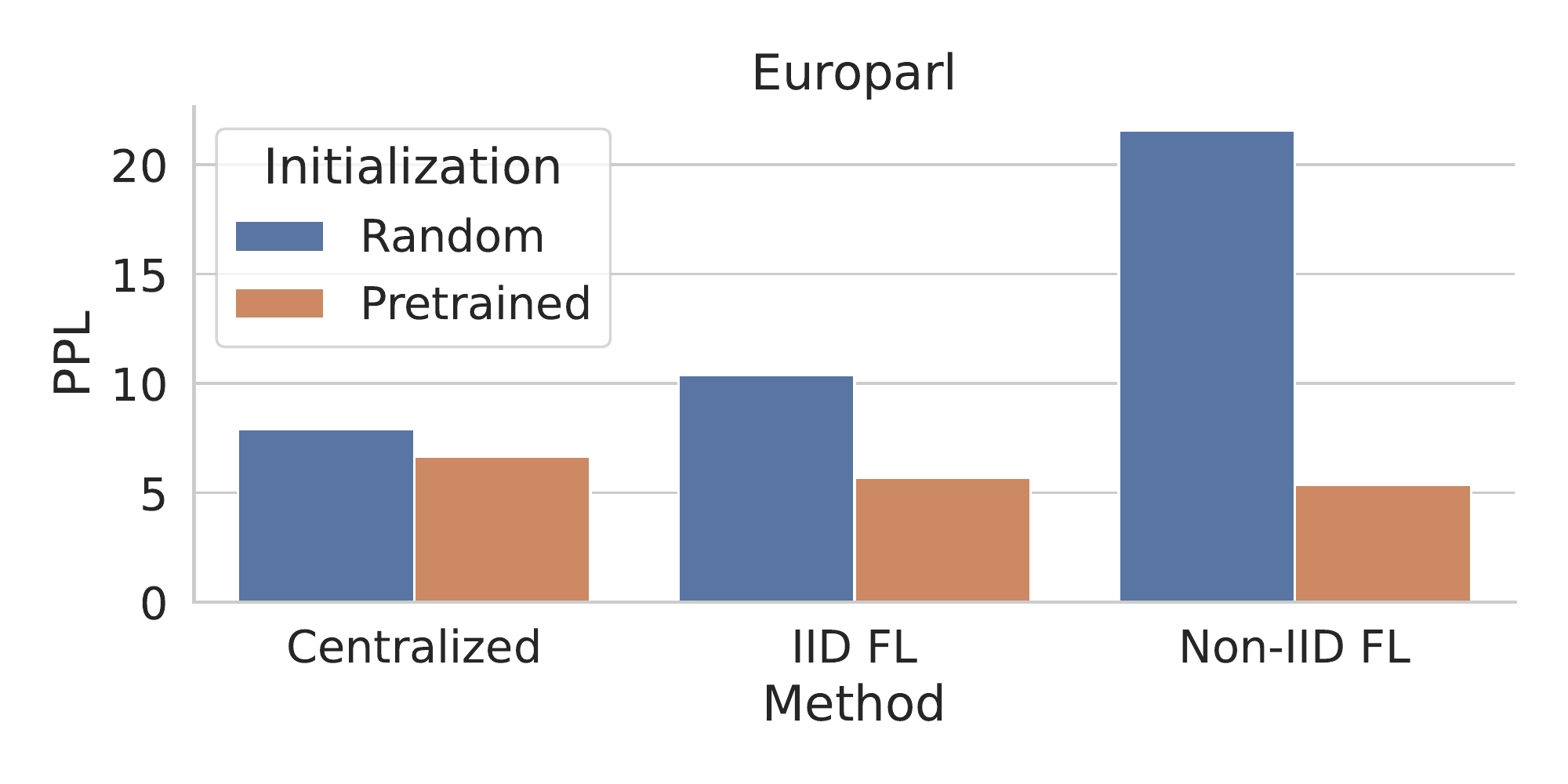}
    \includegraphics[trim=0 10 0 0,clip,width=0.48\textwidth]{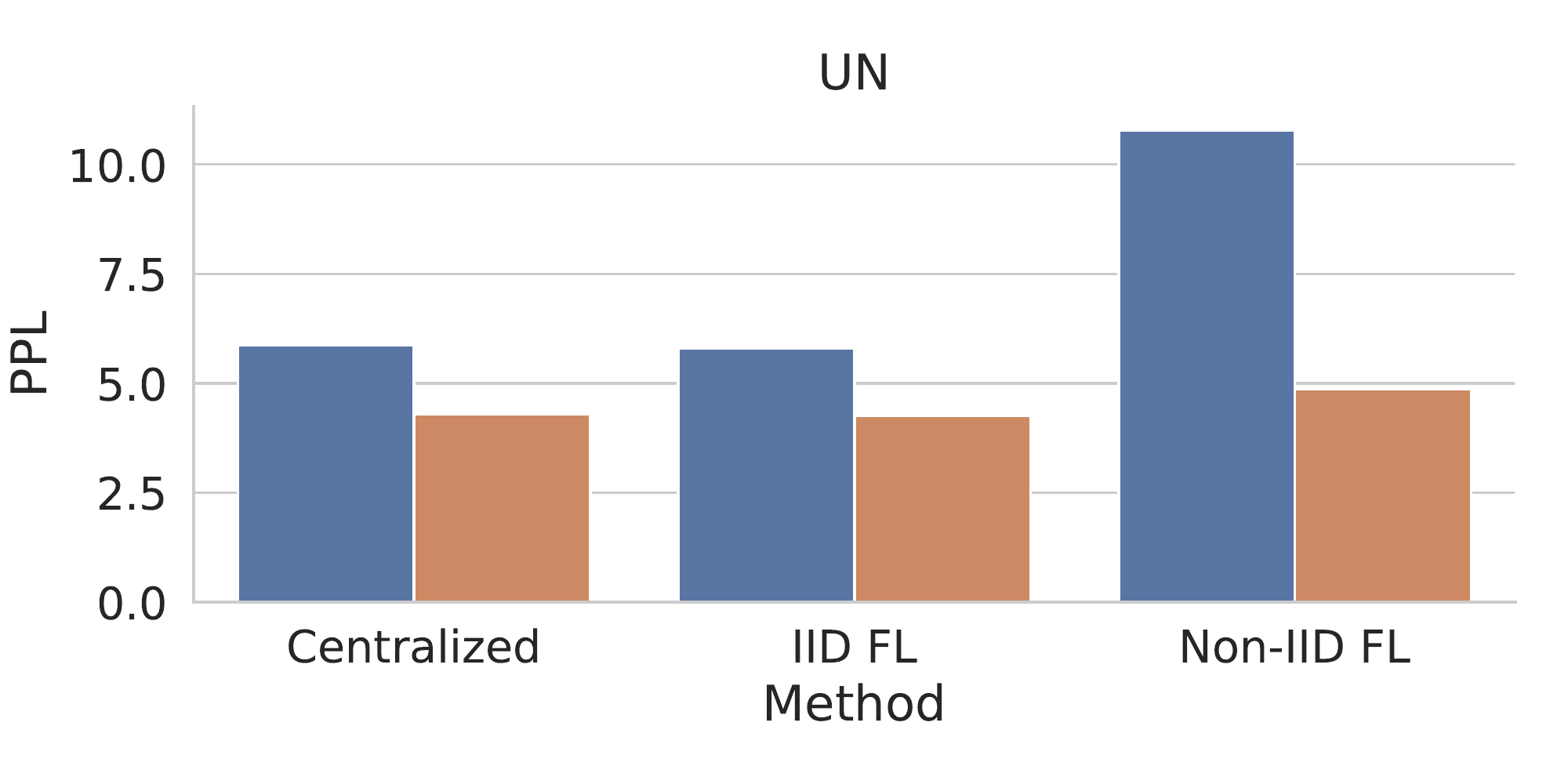}
    \caption{An overview of the language modeling results. Bars indicate the average language perplexity (PPL) over 8 languages for the Europarl dataset and 6 languages for the UN corpus. Lower is better.}
    \label{fig:LM_plot}
\end{figure*}

\begin{table*}[t]
\small
\centering
\begin{tabular}{c|rrrrrrrrr|rrrrrrr}
\toprule
 & \multicolumn{9}{c|}{\textbf{Europarl}} & \multicolumn{7}{c}{\textbf{UN}} \\
 \textbf{M} & \textbf{En} & \textbf{Cs} & \textbf{Lt} & \textbf{Es} & \textbf{Pl}
& \textbf{Fi} & \textbf{Pt} & \textbf{De} & \textbf{Avg} &
\textbf{En} & \textbf{Es} & \textbf{Fr} & \textbf{Ru} & \textbf{Zh} & \textbf{Ar} & \textbf{Avg} \\
\midrule
B & 26.2 & 34.8 & 40.1 & 20.0 & 20.0 & 26.6 & 25.5 & 22.1 & 26.9 & 22.3 & 15.0 & 17.2 & 9.8 & 18.1 & 14.7 & 16.2  \\
\midrule
C & \textbf{19.3} & \textbf{4.5} & \textbf{3.9} & \textbf{8.3} & \textbf{4.7} & \textbf{4.9} & \textbf{7.0} & \textbf{10.8} & \textbf{7.9} & \textbf{9.0} & \textbf{5.2} & \textbf{8.2} & *3.9 & *4.3 & *4.6 & *5.9 \\
I & 26.6 & 5.4 & *4.3 & 11.2 & 5.8 & 5.7 & 8.9 & 15.1 & 10.4 & *9.1 & \textbf{5.2} & *8.4 & \textbf{3.7} & \textbf{3.9} & \textbf{4.5} & \textbf{5.8} \\
N & 50.6 & 7.1 & 11.9 & 16.0 & 17.7 & 12.1 & 35.6 & 21.7 & 21.6 & 12.8 & 11.5 & 14.6 & 9.3 & 8.2 & 8.3 & 10.8 \\
\midrule
C & 12.1 & \textbf{3.7} & \textbf{3.3} & 13.9 & 4.7 & \textbf{4.0} & \textbf{4.8} & *6.8 & 6.7 & *7.0 & *4.1 & \textbf{4.9} & *2.9 & *3.3 & *3.6 & \textbf{4.3} \\
I & 10.5 & *4.0 & 4.2 & *6.1 & \textbf{3.8} & 4.5 & *5.6 & *6.9 & *5.7 & \textbf{6.5} & \textbf{3.9} & 5.7 & \textbf{2.8} & \textbf{3.2} & \textbf{3.5} & \textbf{4.3} \\
N & \textbf{8.8} & \textbf{3.7} & 3.9 & \textbf{6.0} & \textbf{3.8} & *4.4 & *5.6 & \textbf{6.7} & \textbf{5.4} & *7.1 & 4.5 & 6.2 & *3.2 & 4.2 & 4.0 & *4.9  \\
\bottomrule
\end{tabular}
\caption{Results for FL experiments on the LM task. Bold scores indicate the best in the column for the given section. Scores are measured in perplexity (lower is better). The top row (B) is a baseline using the pretrained model with no fine-tuning. The middle rows are trained from \textit{randomly-initialized} models while the bottom rows tune the pretrained model on task data. Due to space we abbreviate: C for Centralized, I for IID FL, and N for non-IID FL.
We sample the mask distribution with 5 seeds and report the mean (standard deviations can be found in the Appendix, Tables \ref{tab:europarl_lm_with_std} and \ref{tab:un_lm_with_std}). Asterisks indicate scores within 2 standard deviations of the best. 
}
\label{tab:lm_results_detailed}
\end{table*}

\subsection{Federated Learning Methods}
We use \algfont{FederatedAveraging} as the primary learning algorithm \citep{mcmahan2017communication}.
\algfont{FederatedAveraging} was introduced alongside the term Federated Learning and has been studied in both learning theory research \cite{stich2019local} and applied work \cite{hard2018federated,fednlp2021}.
In this algorithm, each client runs stochastic gradient descent (SGD) on its local data.
After a specified number of steps, the client transmits its local model to the server, which averages these updates into a single centralized set of parameters. The server then broadcasts the centralized parameters to each client and the process repeats.

\subsection{Client Partitioning}

We consider three different training settings: standard training with no FL (e.g. \textit{centralized} or \textit{C}), FL with IID data (\textit{FL IID} or \textit{I}), where the data for each client is sampled randomly from all data, and FL with non-IID data (\textit{FL non-IID} or \textit{N}) where each client only sees data for one language (or for MT, one direction). See \Cref{fig:task} for a visual depiction of these three client partitioning schemes.

\subsection{Data}
We study three multilingual language tasks, due to their common use in the community: language modeling (LM), machine translation (MT), and text classification (TC). We note that the data we use for training is relatively small; however, this mirrors pratical FL, as each client will not have a large amount of data. We measure scores using perplexity (PPL) for LM,
BLEU \cite{papineni-etal-2002-bleu} for MT, and accuracy for TC.

\paragraph{Europarl} We use the Europarl corpus \citep{koehn2005europarl} taken from transcripts of European Union meetings. We sample data from eight languages: English, Spanish, Portuguese, French, German, Finnish, Polish, Lithuanian, and Czech. We sample 20k of each language for training and 5k for validation/testing, and use it for the LM task.

\paragraph{MTNT} We use the Machine Translation of Noisy Text (MTNT) dataset \cite{michel2018mtnt}, which was the testset for the 2019 WMT robustness challenge.
MTNT was gathered from user comments on Reddit discussion threads and contains noisy text including typos, casual language, and niche terminology. The dataset contains two non-English languages that we use: En $\rightarrow$ Fr and En $\rightarrow$ Ja.
This dataset has been used to test MT systems for robustness to domain shift \citep{li-etal-2019-findings} and is suitable for our experiments since FL deals with client data that is uniquely shifted from centralized data.
For more details on MTNT data preprocessing for M2M-100, see \Cref{app:MTNTM2M100}.

\paragraph{UN Corpus} The UN Corpus \cite{ziemski2016united} consists of official records from the UN proceedings over the years 1990 to 2014, in six languages: English, French, Spanish, Russian, Chinese, and Arabic. We use this data for LM (with 50k instances of training data per language and 5k for validation/testing) as well as three MT directions covering 6 languages (En $\rightarrow$ Fr, Ar $\rightarrow$ Es, Ru $\rightarrow$ Zh). Following previous work in MT adaption (see MTNT above) we sample 10k in each direction for training and 5k each for evaluation sets.

\paragraph{NC Corpus}
For text classification we use the News Classification (NC) dataset from the XGLUE benchmark for cross-lingual language understanding \citep{Liang2020XGLUEAN}. This is a classification problem with 10 classes across 5 languages: English, Spanish, French, German, and Russian. We predict the article category given the article title and body (e.g. finance, sports, travel). Since only 10k annotated examples are available for each language (excluding the official test set), we sample 8k instances for training and 1k for evaluation sets. Note that although XGLUE is made for cross-lingual evaluation, we use it for multilingual evaluation.

\newcommand*{\thead}[1]{\multicolumn{1}{c}{\bfseries #1}}

\begin{table*}[ht]
\centering
\begin{tabular}{l|rrr|rrrr}
\toprule
 & \multicolumn{3}{c|}{\textbf{MTNT}} & \multicolumn{4}{c}{\textbf{UN}} \\
 \textbf{Method} & \textbf{En-Fr} & \textbf{En-Ja}  & \textbf{Avg} &
\textbf{En-Fr} & \textbf{Ar-Es} & \textbf{Ru-Zh} & \textbf{Avg} \\
\midrule
No Training & 30.7 & 14.1 & 22.4 & 31.4 & 27.4 & 27.9 & 28.9 \\
\midrule
Centralized & 31.8 & *15.4 & 23.6 & 37.3 & 35.9 & 34.1 & 35.8 \\
IID FL & \textbf{33.1} & \textbf{15.6} & \textbf{24.4} & \textbf{38.6} & \textbf{36.9} & *35.6 & \textbf{37.0} \\
non-IID FL & *32.9 & \textbf{15.6} & 24.3 & 37.9 & *36.6 & \textbf{35.7} & 36.7 \\
\bottomrule
\end{tabular}
\caption{Results for FL experiments on the Machine Translation task. Bold scores indicate the best in the column, while asterisks indicate scores that are statistically similar to the best according to a paired bootstrap resampling test. Scores are measured with \algfont{sacreBLEU} \cite{post2018call}, higher is better.}
\label{tab:mt_results_detailed}
\end{table*}

\begin{table*}[ht]
\centering
\begin{tabular}{l|rrrrrr}
\toprule
 \thead{Method} & \thead{En} & \thead{Es}  & \thead{Fr} & \thead{De} & \thead{Ru} & \thead{Avg}  \\
\midrule
Centralized & 86.6 \textpm\ 0.3 & 77.5 \textpm\ 1.2 & 74.9 \textpm\ 1.6 & *82.3 \textpm\ 1.6 & 80.7 \textpm\ 0.7 & 80.4 \textpm\ 0.6 \\
IID FL & \textbf{88.0} \textpm\ 0.6 & \textbf{79.8} \textpm\ 0.5 & \textbf{76.4} \textpm\ 0.6 & \textbf{82.6} \textpm\ 0.6 & \textbf{82.5} \textpm\ 0.4 & \textbf{81.8} \textpm\ 0.3 \\
non-IID FL & 81.0 \textpm\ 0.9 & 69.3 \textpm\ 1.6 & 73.7 \textpm\ 1.0 & 76.0 \textpm\ 0.3 & 71.9 \textpm\ 1.1 & 74.4 \textpm\ 0.5  \\
\midrule
Centralized & 93.5 \textpm\ 0.7 & *86.3 \textpm\ 0.5 & \textbf{82.9} \textpm\ 0.3 & \textbf{89.6} \textpm\ 0.1 & *88.5 \textpm\ 0.4 & *88.1 \textpm\ 0.2 \\
IID FL & \textbf{94.0} \textpm\ 0.2 & \textbf{86.9} \textpm\ 1.1 & 82.1 \textpm\ 0.7 & \textbf{89.6} \textpm\ 0.2 & \textbf{89.1} \textpm\ 1.2 & \textbf{88.3} \textpm\ 0.3  \\
non-IID FL & 92.5 \textpm\ 0.1 & *86.1 \textpm\ 0.6 & 81.4 \textpm\ 0.3 & 88.8 \textpm\ 0.1 & 84.5 \textpm\ 0.7 & 86.7 \textpm\ 0.1   \\
\bottomrule
\end{tabular}
\caption{Results for FL experiments on the Text Classification task. Bold scores indicate the best in the column, while asterisks indicate scores within two standard deviations of the best. Scores are the mean of training with 3 different seeds, $\pm$ denotes the standard deviation.  Scores are measured with accuracy, higher is better. The top rows are trained from random initialization while the bottom rows initialize from the pretrained model.}
\label{tab:tc_results_detailed}
\end{table*}

\subsection{Modeling}
For language modeling and text classification, we examine two different initialization settings: (1) fine-tuning from a pretrained multilingual model or (2) training the same multilingual model architecture but doing so with randomly initialized weights. For the MT experiments, we omit the \textit{randomly-initialized} results as MT systems generally need large amounts of data to produce good results (see \Cref{app:mt_randominit} for more details).

Our base model for the LM task is a distilled version of the mBERT model (134M parameters), shown to perform well across many languages \citep{sanh2019distilbert,devlin2019bert} while being smaller than the full mBERT.\footnote{We note that mBERT uses masked language modeling (MLM) instead of standard language modeling, however, for the purposes of our analysis (as we do not seek to compare direct scores to previous work) MLM suffices. Furthermore, most multilingual PLMs train via some version of MLM.} For MT, we use the M2M-100 model \citep{Fan2020BeyondEM} with 418M parameters, a many-to-many MT model that can translate between any pairing of 100 languages. For text classification, we use the XLM-RoBERTa base sized model (270M parameters).
We note that although there are other PLMs to consider, we focus on testing a varied set of commonly used, high-performing PLMs.

\subsection{Training}
We use the Flower framework \cite{beutel2020flower} for federated training and evaluation due to its ease of use and strong community support. We use Hugging Face's \textit{transformers} library \cite{wolf2019huggingface} for loading pretrained models and PyTorch as the underlying differentiation framework \cite{NEURIPS2019_9015}. We train each LM model for 100 epochs if pretrained or 200 epochs if randomly initialized. For MT, we train for 25 epochs and for TC we train for 10 epochs if pretrained and 50 epochs if randomly initialized. For other hyperparameters and compute settings, see \Cref{app:hyperparameters}.

\section{Results}

\paragraph{Language Modeling}
In \Cref{fig:LM_plot} we see the overall results of the language modeling task across the two datasets. As expected, the randomly initialized models perform much worse than the pretrained models. The gap between between FL and centralized methods is smaller when using pretrained models, indicating that pretrained models are an effective initialization for federated learning.

In \Cref{tab:lm_results_detailed} we show results broken down by language. Since the fine-tuning task is the same as the pretraining objective (masked language modeling), we can use the pretrained model as a baseline (top row, B).   
In the randomly initialized category, the centralized model is the same or better than the FL methods in every single language, across both datasets.
In the pretrained section the results are more mixed, with the centralized model winning or tying in 5 of the 8 Europarl languages and obtaining similar scores on the UN corpus.
We also see that the randomly initialized non-IID model appears to diverge for some of the Europarl languages. 

Examining the difference between IID FL and non-IID FL, we see that IID FL performs better on average in three of the four settings. However, when initializing with a pretrained model, the performance gap narrows.

\paragraph{Machine Translation}
\Cref{tab:mt_results_detailed} exhibits results on tuning a machine translation model on a domain specific dataset. We see that on the MTNT dataset, both FL algorithms actually outperform centralized learning (24.4 avg. BLEU for IID FL vs 23.6 for Centralized). The scores on Japanese are very similar for all models, possibly reflecting the difficulty of the task. On the UN corpus, we see again that the IID FL model performs best.

Since the fine-tuning task matches the original M2M-100 task, we can use the pretrained model directly as a baseline. In all cases, fine-tuning shows an improvement (first row, No Training baseline). Note that our scores are not directly comparable to other work as we use a smaller training set.

\paragraph{Text Classification}
\Cref{tab:tc_results_detailed} shows results on text classification. We see that when initialized randomly, non-IID FL shows a large drop in performance compared to the two other methods (i.e. more than 5 points worse than the Centralized method).
Initializing with the pretrained model yields a modest though consistent improvement for all three models (80.4\% accuracy vs 88.3\% accuracy for Centralized).\footnote{Note that although the setups are not the same (e.g. XGLUE is cross-lingual rather than multilingual) our scores are slightly higher than those reported in the original paper.} Furthermore, with a pretrained initialization the non-IID FL method scores become significantly closer to the other two methods, with less than a two point difference between them (86.7\% non-IID FL vs 88.3\% IID FL).

\paragraph{Discussion}
Our examination of multilingual FL indicates that performance is similar when pretrained models are used.
Despite the fact that local models are averaged together, non-IID data partitioning (where each client sees only one language) has only a small impact on final multilingual performance, when using pretrained models.
These findings suggest that, when possible, practitioners who need multilingual federated learning should employ pretrained models in order to gain the privacy benefits of federated learning, without taking much (if any) of a performance loss to do so.

In several cases, we found that IID FL or non-IID FL could even outperform centralized learning. We leave investigation of this phenomena for future work but note a couple of possible explanations.
First, FL with \algfont{FederatedAveraging} may have similar implicit regularization effects to checkpoint averaging, a common technique when using transformer models (noted in \citealt{vaswani2017attention}, \citealt{edunov-etal-2018-understanding}, etc.).
Furthermore, there may be other regularization effects during federated fine-tuning, as transformer training is known to be unstable and sensitive to optimization choices (\citealt{mosbach2020stability}, \citealt{nguyen2019transformers}).

Overall, our analysis shows that our conclusions hold for different multilingual models, on disparate NLP tasks, and across 13 different languages. We acknowledge that the languages used in this study are generally considered higher-resource, but expect that these conclusions will continue to hold as long as the pretrained model is effective on the target language (or language pairs, for MT).

\section{Conclusion}
In this work we provided the first analysis of multilingual language data on federated learning algorithms.
We found that fine-tuning a pretrained model with FL methods can yield similar performance to centralized learning, even when clients are partitioned by language (non-IID FL).

However, models trained from random initializations still show a large gap between centralized and federated learning.
Our results suggest that learning on private partitioned data is possible without having to incur a large performance penalty. 
We hope that these results will aid practitioners in using FL (and also downstream consumers) and inspire the broader community to consider multilingual data in future federated learning research for natural language processing.

\bibliography{anthology,custom}
\bibliographystyle{acl_natbib}



\appendix

\section{Hyperparameters}
\label{app:hyperparameters}
Each LM experiment ran for approximately a day each on a 6 GPU cluster of RTX 6000 GPUs with 24GB of memory per GPU. The MT experiments took approximately 12 hours each and the TC experiments took around 3 hours each, all on the same cluster.

We use the AdamW optimizer \cite{loshchilov2017decoupled,kingma2014adam} for all experiments (shown to be effective for FL in \citealt{fednlp2021}). Each client goes through a full epoch of local learning before synchronizing with the server.

For MT, we report results using the 5e-5 learning rate, as we found in initial results (as have others also, see Appendix B of \citet{stickland2020recipes} as one example) that MT experiments are generally consistent over learning rates when fine-tuning.
For language modeling and text classification, we use three different learning rates (1e-4, 5e-5, 1e-5). All models were selected using the best performing version on the validation set, for the given model and training setting.
For both tasks, we use early stopping (5 epochs of no improvement for MT and TC, 10 epochs for LM).

We use the standard \algfont{sacreBLEU} settings: nrefs:1, mixed case, eff:no, tok:13a, smooth:exp, and version 2.0.0. For Ja and Zh we use their respective tokenizers.

\section{Randomly Initialized MT}
\label{app:mt_randominit}
We do not report results for randomly initialized training of MT systems, as large neural MT systems generally need large amounts of data to be effective. We ran experiments for the MTNT dataset from random initializations, running for twice as many epochs. Resulting models appeared to converge by loss but had extremely low BLEU scores. Thus, we only include pretrained results in \Cref{tab:mt_results_detailed}.

\section{MTNT Data Preprocessing for M2M-100}
\label{app:MTNTM2M100}
M2M-100 was trained using scripts that removed input with ``excess punctuation." We follow this in preparing MTNT training data. We use all En $\rightarrow$ Ja data (consisting of approximately 6k instances) and take the corresponding En $\rightarrow$ Fr instances, randomly sampling additional instances until there are the same number of instances in each direction. We sample an equal number of training instances as we are testing the effects of multilingual data, rather than unequal dataset sizes. We then remove the training instances with excess punctuation (or sentences less than 3 characters) following the M2M-100 script. This leaves 5605 instances in each direction for training. We use the standard MTNT dev and test sets, as-is, consisting of approximately 1k data points.

\section{Full LM Results}
\label{app:FullLM}
We show the full results of the LM experiments, with standard deviations over five random seeds in Tables \ref{tab:europarl_lm_with_std} and \ref{tab:un_lm_with_std}.

\begin{table*}[t]
\small
\centering
\begin{tabular}{c|rrrrrrrrr}
\toprule
 & \multicolumn{9}{c}{\textbf{Europarl}} \\
 \thead{M} & \thead{En} & \thead{Cs} & \thead{Lt} & \thead{Es} & \thead{Pl}
& \thead{Fi} & \thead{Pt} & \thead{De} & \thead{Avg} \\
\midrule
B & 26.2 \textpm\ 2.4 & 34.8 \textpm\ 1.8 & 40.1 \textpm\ 2.3 & 20.0 \textpm\ 1.3 & 20.0 \textpm\ 1.3 & 26.6 \textpm\ 1.4 & 25.5 \textpm\ 2.0 & 22.1 \textpm\ 1.8 & 26.9 \textpm\ 1.8 \\\midrule
C & 19.3 \textpm\ 1.5 & 4.5 \textpm\ 0.4 & 3.9 \textpm\ 0.3 & 8.3 \textpm\ 0.7 & 4.7 \textpm\ 0.3 & 4.9 \textpm\ 0.3 & 7.0 \textpm\ 0.6 & 10.8 \textpm\ 0.8 & 7.9 \textpm\ 0.6 \\
I & 26.6 \textpm\ 1.7 & 5.4 \textpm\ 0.4 & 4.3 \textpm\ 0.3 & 11.2 \textpm\ 0.9 & 5.8 \textpm\ 0.4 & 5.7 \textpm\ 0.3 & 8.9 \textpm\ 0.7 & 15.1 \textpm\ 1.1 & 10.4 \textpm\ 0.7 \\
N & 50.6 \textpm\ 2.9 & 7.1 \textpm\ 0.5 & 11.9 \textpm\ 0.9 & 16.0 \textpm\ 1.2 & 17.7 \textpm\ 1.2 & 12.1 \textpm\ 0.7 & 35.6 \textpm\ 2.8 & 21.7 \textpm\ 1.4 & 21.6 \textpm\ 1.4  \\
\midrule
C & 12.1 \textpm\ 0.9 & 3.7 \textpm\ 0.3 & 3.3 \textpm\ 0.2 & 13.9 \textpm\ 1.2 & 4.7 \textpm\ 0.4 & 4.0 \textpm\ 0.2 & 4.8 \textpm\ 0.4 & 6.8 \textpm\ 0.6 & 6.7 \textpm\ 0.5 \\
I & 10.5 \textpm\ 0.9 & 4.0 \textpm\ 0.3 & 4.2 \textpm\ 0.3 & 6.1 \textpm\ 0.5 & 3.8 \textpm\ 0.3 & 4.5 \textpm\ 0.3 & 5.6 \textpm\ 0.4 & 6.9 \textpm\ 0.5 & 5.7 \textpm\ 0.4  \\
N & 8.8 \textpm\ 0.6 & 3.7 \textpm\ 0.3 & 3.9 \textpm\ 0.3 & 6.0 \textpm\ 0.5 & 3.8 \textpm\ 0.3 & 4.4 \textpm\ 0.3 & 5.6 \textpm\ 0.4 & 6.7 \textpm\ 0.5 & 5.4 \textpm\ 0.4  \\
\bottomrule
\end{tabular}
\caption{Results for the LM FL experiments on the Europarl Corpus. Bold scores indicate the best in the column for the given section. Scores are measured in perplexity (lower is better). The top row (B) is a baseline using the pretrained model with no further tuning. The middle rows are trained from \textit{randomly-initialized} models while the bottom rows tune the pretrained model on task data. Due to space we abbreviate: C for Centralized, I for IID FL, and N for non-IID FL.
}
\label{tab:europarl_lm_with_std}
\end{table*}

\begin{table*}[t]
\small
\centering
\begin{tabular}{c|rrrrrrr}
\toprule
 & \multicolumn{7}{c}{\textbf{UN}} \\
 \thead{M} & \thead{En} & \thead{Es} & \thead{Fr} & \thead{Ru} & \thead{Zh} & \thead{Ar} & \thead{Avg} \\
  \midrule
 B &  22.3 \textpm\ 2.0 & 15.0 \textpm\ 0.9 & 17.2 \textpm\ 1.1 & 9.8 \textpm\ 0.6 & 18.1 \textpm\ 1.0 & 14.7 \textpm\ 0.8 & 16.2 \textpm\ 1.0 \\
\midrule
C & 9.0 \textpm\ 0.6 & 5.2 \textpm\ 0.3 & 8.2 \textpm\ 0.5 & 3.9 \textpm\ 0.3 & 4.3 \textpm\ 0.2 & 4.6 \textpm\ 0.3 & 5.9 \textpm\ 0.4 \\
I &  9.1 \textpm\ 0.7 & 5.2 \textpm\ 0.3 & 8.4 \textpm\ 0.4 & 3.7 \textpm\ 0.3 & 3.9 \textpm\ 0.2 & 4.5 \textpm\ 0.2 & 5.8 \textpm\ 0.3 \\
N & 12.8 \textpm\ 0.9 & 11.5 \textpm\ 0.7 & 14.6 \textpm\ 0.8 & 9.3 \textpm\ 0.7 & 8.2 \textpm\ 0.5 & 8.3 \textpm\ 0.4 & 10.8 \textpm\ 0.6 \\
\midrule
C & 7.0 \textpm\ 0.5 & 4.1 \textpm\ 0.2 & 4.9 \textpm\ 0.3 & 2.9 \textpm\ 0.2 & 3.3 \textpm\ 0.2 & 3.6 \textpm\ 0.2 & 4.3 \textpm\ 0.3  \\
I & 6.5 \textpm\ 0.5 & 3.9 \textpm\ 0.2 & 5.7 \textpm\ 0.3 & 2.8 \textpm\ 0.2 & 3.2 \textpm\ 0.2 & 3.5 \textpm\ 0.2 & 4.3 \textpm\ 0.3 \\
N & 7.1 \textpm\ 0.5 & 4.5 \textpm\ 0.3 & 6.2 \textpm\ 0.3 & 3.2 \textpm\ 0.2 & 4.2 \textpm\ 0.2 & 4.0 \textpm\ 0.2 & 4.9 \textpm\ 0.3 \\
\bottomrule
\end{tabular}
\caption{Results for the LM FL experiments on the UN Corpus. Bold scores indicate the best in the column for the given section. Scores are measured in perplexity (lower is better). The top row (B) is a baseline using the pretrained model with no further tuning. The middle rows are trained from \textit{randomly-initialized} models while the bottom rows tune the pretrained model on task data. Due to space we abbreviate: C for Centralized, I for IID FL, and N for non-IID FL.}
\label{tab:un_lm_with_std}
\end{table*}

\end{document}